\documentclass[11pt,a4paper]{article}
\usepackage[hyperref]{conll-2019}
\usepackage{times}
\usepackage{latexsym}

\usepackage{url}

\usepackage{graphicx}
\usepackage{multirow}
\usepackage{lipsum}
\usepackage{amsmath}
\usepackage{amssymb}

\aclfinalcopy 

\title{A Robust Data-Driven Approach for Dialogue State Tracking of \\Unseen Slot Values}

\author{Vevake Balaraman$^{1,2}$, Bernardo Magnini$^1$\\
  $^1$Fondazione Bruno Kessler, Trento, Italy \\
  $^2$ICT Doctoral School, University of Trento, Italy \\
{\tt\{balaraman, magnini\}@fbk.eu} \\}
\date{}

\begin{document}
\maketitle
\begin{abstract}
A Dialogue State Tracker is a key component in dialogue systems which estimates the beliefs of possible user goals at each dialogue turn.
Deep learning approaches using recurrent neural networks have shown state-of-the-art performance for the task of dialogue state tracking.
Generally, these approaches assume a predefined candidate list and struggle to predict any new dialogue state values that are not seen during training.
This makes extending the candidate list for a slot without model retaining infeasible and also has limitations in modelling for low resource domains where training data for slot values are expensive.
In this paper, we propose a novel dialogue state tracker based on copying mechanism that can effectively track such unseen slot values without compromising performance on slot values seen during training.
The proposed model is also flexible in extending the candidate list without requiring any retraining or change in the model.
We evaluate the proposed model on various benchmark datasets (DSTC2, DSTC3 and WoZ2.0) and show that our approach, outperform other end-to-end data-driven approaches in tracking unseen slot values and also provides significant advantages in modelling for DST.
\end{abstract}

\section{Introduction}
Spoken dialogue systems (SDS), or conversational systems, are designed to interact and assist users using speech and natural language to achieve a goal \cite{Henderson2015}.
A dialogue state tracker (DST) is a crucial component of the dialogue system that is responsible for tracking the user goals as slot-value pairs during the conversation.
The goals tracked at any given turn in the dialogue is referred to as the \textit{dialog state}.
This dialog state is then used by the dialog manager (DM) to decide on the next action.
Though end-to-end (E2E) approaches for dialogue systems has attracted recent research, dialog state tracking still remains an integral part in those systems \cite{Liu2017, Wen2017, Li2017} due to the fact that dialogue systems always needs to interact with backend knowledge base (KB) using the dialogue state tracked by DST.  

\begin{figure}
  \centering
  \includegraphics[width=\linewidth]{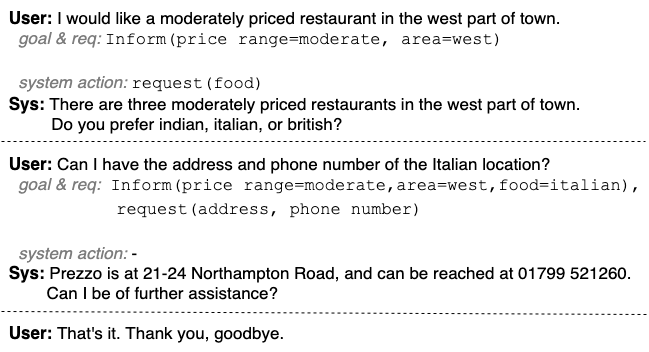}
  \caption{An annotated dialogue from the WoZ2.0 dataset, with each turn separated by a dashed line.}
  \label{fig:example}
  \vspace{-4mm}
\end{figure}

The dialog state is typically represented as a probability distribution over the possible states.
Given a set of slots $S$ and a set of possible values $V_s$ for each slot $s \in S$, the dialogue state estimates the distribution over the values $V_s$ for each slot $s$ based on the user utterance and the dialogue history.
The Figure \ref{fig:example} shows a sample conversation with the dialog states at each turn.

Deep neural network techniques such as recurrent neural network and convolutional neural networks are the current state-of-the-art models for DST \cite{Henderson2014, Wen2017, NBT, Ren2018, GLAD}.
Traditionally, the DST algorithms rely on a predefined \textit{domain ontology} that describes a fixed candidate list for each possible slot.
The DST's are trained to output distribution over these predefined slot value set and lack in flexibility to effectively predict for new values of the existing slots.
This is an important aspect of DST as in real-world scenario, the slot values in the ontology requires frequent updates \cite{Henderson2014RobustDS, Xu2018}.
For example in the domain of restaurant booking, a new cuisine could be added for the slot $food$ and the DST should be able to track this new slot without any retraining or change in model as it has already been trained for other such cuisines initially.
Moreover, even if the new slot value was anticipated and added to the ontology, it would still require training data to train and predict for that model.
This limitation is of high importance in low resource domains where obtaining data for all slot values is expensive.


The new slot values not seen in training referred to as \textit{unseen slot values} are often handled by DST either using rule-based approaches or delexicalization techniques (replacing slot values in input with a generic token) \cite{dstc3challenge, Henderson2014, Wen2017}.
While end-to-end approaches for DST has shown state-of-the-art performance in predicting \textit{dialog state} from predefined slot value set (\textit{seen slot values}) \cite{NBT, GLAD}, they suffer from transferring this knowledge to track any new slot values.
This is due to the fact that such models are tailored to learn only for the slot values for which training data is available.
Any update for the slot values requires new training data for that value and model retraining.
While approaches such as \cite{Yang2018, yoshida2018, Xu2018} has tried to address this using data-driven approaches, they often make compromise in the performance of seen slot values to accommodate for the unseen slot values.
Their evaluation shows that a large performance gap still persists in predicting for seen and unseen slot values.

In this paper, we focus on this important problem of tracking unseen slot values in DST which has so far relied on using hand crafted features to handle them effectively.
In particular, we investigate on copying mechanism to build a robust DST that can effectively track both seen and unseen slot values without requiring any hand crafted features.
Our experiments show that in-addition to learning the presence of a value in the input, the model can also learn the context at which it occur and use this information in predicting for the unseen slot values making it suitable for low resource domains as well. 

The major contributions of this paper are as follows: i)we address a real-world problem in DST which is predicting for unseen slot values using a data-driven approach without any hand-crafted features, 
ii)we present a novel E2E approach based on copying mechanism for DST,
iii)we conduct extensive experimental study on various benchmark datasets of DST showing the effectiveness of the approach and
iv)the implementation of our approach is made publicly available\footnote{\textit{will be made available in final version}} for further research.

\section{Related Work}
Traditional dialogue systems consist of a spoken language understanding (SLU) component that performs slot-filling to detect slot-value pairs expressed in the input.
The dialog state tracker then uses this information together with the past dialogue context and updates it belief \cite{Wang2013, Sun2015}.
This approach has the limitation of error propagation from the SLU model to DST, and hence the recent research has been focused on jointly modelling the SLU and DST \cite{Henderson2014, Wen2017}.
\cite{Henderson2014} proposed a word-based DST that jointly models SLU and DST, and directly maps from the utterances to an updated belief state.
This approach relies on a semantic lexicon for the delexicalization to identify mentions of unseen slot values.
\cite{NBT} proposed a data-driven approach for DST named neural belied tracker(NBT) which learns a vector representation for each slot-value pair and compares it with the vector representation of the user utterance to predict if the user has expressed the corresponding slot-value pair.

Global-Locally Self-Attentive Dialogue State Tracker (GLAD) was proposed by \cite{GLAD} which models a single global bidirectional-LSTM (BiLSTM) for all slots and a local BiLSTM for each slot.
This global and local representations are then combined using attention which then is used by a scoring module to obtain scores for each slot-value pair.
Since also the slot values are encoded by the BiLSTM, this is ineffective for predicting on unseen slot-values.
The StateNet proposed by \cite{Ren2018} uses LSTM network to create a vector representation of the user utterance which is then compared against the vector representation of the slot-value candidates.
This approach is extendable to new slot-value candidates provided they have a pre-trained semantic embedding vector available.
This forces a constraint on the vocabulary of the slot values.

The most relevant works addressing the problem of unseen slot-values are \cite{Xu2018, Yang2018, Jang2016, yoshida2018}.
\cite{Xu2018} proposed a joint model for DST based on pointer networks which outputs a distribution over the tokens in user utterance.
This approach assumes no candidate set for the values and since it can only output from the input vocabulary, a semantic lexicon is required for post-processing of output to normalize for the value.
\cite{Yang2018} proposed an hierarchical dialog state tracker (HDSTM) that consists of a detection module to identify if an unseen slot value occurs in the user utterance.
Based on the detection module output, the belief state is updated using an update mechanism.
Cosine similarity based DSTs were proposed by \cite{Jang2016, yoshida2018} based on the attention mechanism.
The attention weights are used to obtain a weighted sum of input word embedding vectors which is used to estimate the probability using  with each value.
Since the model prediction is largely based on the similarity of the word vectors, the model struggles to learn from other features available in the input source.
\cite{Rastogi2017} proposed a a scalable multi-domain DST to address for slots for which values set is unbounded.
However their system relies on an SLU to generate the candidates for a slot.

\section{Copying Mechanism}
In this section we briefly describe the copying mechanism introduced in \large{C}\small{OPY}\large{N}\small{ET} \normalsize \cite{Gu2016}, which is the core of our model for tracking unseen slot values.
The copying mechanism refers to the capacity to refer or copy certain segments in the input sentence into the output.
This is a natural behaviour in human language communication, where the response to a certain input sentence may contain segments from the input itself. For example, in \textit{extractive} text summarization \cite{Gupta2010}, the model needs to extract sentences or phrases from an input document and to combine them into an output summary.
The \large{C}\small{OPY}\large{N}\small{ET} \normalsize model builds on top of the encoder-decoder architecture of \cite{Bahdanau2015}, by adding a copying mechanism so as to copy certain relevant input tokens into the output.

In a general sequence-to-sequence model, the encoder takes the source sentence as input $x$ and iteratively processes it to produce hidden states $h_t$ at each timestep $t$ \cite{sutskever2014}.
The decoder then takes the encoder hidden states $h$ and uses an attention mechanism to weight these hidden states based on the previous hidden state of the decoder $s_{t-1}$ to generate a probability distribution (using \textit{softmax} activation function) over output tokens at timestep $t$ of the decoder.
Since this approach relies on generating output tokens only from the predefined output vocabulary, any new tokens in the source sentence can not be generated by the model.

The \large{C}\small{OPY}\large{N}\small{ET} \normalsize mechanism builds on the same model, but the probability of generating any target word $y_t$ is given by the combination of two probabilities, namely generate-mode and copy-mode.
The score for the generate-mode is the same as the generic decoder, while the score for the copy-mode $\psi_c$ is calculated for each word in the source input $x_j$ as follows:
\begin{equation}
    \psi_c(y_t = x_j) = \sigma(h_j^T W)s_t
\end{equation}

where $h_j$ is the hidden state for input $x_j$, $W$ is the parameter learned during training and $s_t$ is the hidden state of decoder at time $t$.
The normalization of this copy score gives us a probability distribution over tokens present in the input $X$.
Finally the distributions for generating the token and for copying the token from the input token are combined to yield the final distribution.

The details on how this copying mechanism is adapted for the task of dialogue state tracking are explained in the next section.

\section{Proposed Model}
The proposed DST is designed to predict for slot value pairs for a given turn in the dialogue and be flexible to slot value updates in the ontology.
For a dialogue turn, given the user utterance $U$, the previous system action $A$ and the value set $V_s$ for slot $s\in S$, the proposed model provides a binary prediction for each value $v \in V_s$.
\begin{equation}
    P_s = DST(U, A, V_s)
\end{equation}

The proposed model is based on encoder-decoder architecture.
The encoder module consists of a recurrent neural network (RNN) that takes as input the user utterance $U$ and the previous system action $A$, and outputs a vector representation $h$.
The decoder then receives the representation $h$ and the slot-values $v \in V_s$ and estimates a score for each slot-value.
The scores are estimated in two stages namely \textit{value score} and \textit{copy score}.
\textit{Value score} helps the model in predicting the seen slot-values while copy score helps the model in predicting the unseen slot-values.
Combining both these score in a single model helps us build a robust DST.
We implement a shared encoder, the outputs of which is processed by multiple decoders (one for each slot $s$) to generate output predictions.

\subsection{Encoder}
The encoder provides a representation for each token in the input and a context vector that ideally summarizes the input.
The encoder takes in the previous system action $A$ and the current user utterance $U$ as inputs and processes them iteratively to output a hidden representation for each token in the input as well as the context vector.

Let the user utterance at time $t$ be denoted as $U=\{u_1,u_2,...,u_k\}$ with $k$ words and $A$ denote the previous system action.
The system action $A$ is converted into a sequence of tokens that include the $action, slot$ and $value$ (e.g. \textit{confirm(food=italian)} $\rightarrow$ confirm food Italian) and is denoted as $A=\{a_1, a_2,..,a_l\}$.
In case of multiple actions expressed by the system, we concatenate them together.
The user utterance $U$ and system action $A$ are then concatenated forming the input $X$ to the encoder.
\[X = [a_1,...a_l;u_1,..u_k] = [x_1,x_2,...x_n]\]
where $[\:;\:]$ denotes concatenation.
Each input tokens in $\{x_1,x_2,..x_n\}$ are then represented as a vector $\{\boldsymbol{x_1},\boldsymbol{x_2},..,\boldsymbol{x_n}\}$ by an embedding matrix $E \in \mathbb{R}^{|v|\times d}$ where $|v|$ is the vocabulary size and $d$ is the embedding dimension.
This representation is then input to the bidirectional-LSTM \cite{hochreiter1997} that processes the input in both forward and backward directions to yield the hidden representation as follows,
\begin{align}
    \overrightarrow{h}_t &= LSTM_f(\overrightarrow{h}_{t-1}, \boldsymbol{x_t}) \\
    \overleftarrow{h}_t &= LSTM_b(\overleftarrow{h}_{t+1}, \boldsymbol{x_t})
\end{align}
where $LSTM_f(.)$, $LSTM_b(.)$ are forward and backward LSTMs. $\overrightarrow{h}_t$ and $\overleftarrow{h}_t$ are the corresponding hidden states of forward and backward LSTMs at time $t$.
The representations $h_t$ for each token in the input and the overall input representation $h_L$ are then obtained as follows,
\begin{align}
    h_t &= [\overrightarrow{h}_t;\overleftarrow{h}_t] \\
    h_L & = [\overrightarrow{h}_n;\overleftarrow{h}_1]
\end{align}

Since our model uses a shared encoder for all slots, the outputs of the encoder $h_t$ and $h_L$ are used by slot specific decoders for prediction on corresponding slots.

\subsection{Decoder}
The decoder of the model is a classifier that predicts the probability for each of the possible values $v \in V_s$  of a given slot $s \in S$.
It takes in the hidden representations $h_t$ and $h_L$ of the encoder, and the set of possible values $V_s$ for a given slot $s$ and computes the probability for each value being expressed as a constraint.
Initially, each of the slot-values are represented by a vector $\boldsymbol{v}$ using the same embedding matrix $E$ of the encoder.
For slot-values with multiple tokens, their corresponding embeddings are summed together to yield a single vector.
The embeddings are then transformed as following to obtain a representation of the slot values.
\begin{equation}
    Z_s = W_s\boldsymbol{V}_s^T \\
\end{equation}
where $\boldsymbol{V}_s = \{\boldsymbol{v_1}, \boldsymbol{v_2},..\}$ for slot $s$ and $W_s$ is the parameter learned during training.
The encoder hidden state $h_L$ is then transformed to obtain a slot specific representation of the user utterance as follows,
\begin{equation}
    S = tanh(W_hh_L)
\end{equation}

The slot-value representations $Z_s$, slot specific input representation $S$ and the hidden states of the input tokens $h_t$ are then used to obtain the probability for each slot-value $p(slot=v)$.
\begin{equation}
    p(slot=v) = \sigma(\psi_p(v) + \psi_c(v)) \\
\end{equation}
where $\psi_p$ is the value score and $\psi_c$ is the copy score.

\subsubsection{Copy score}
To address the ability of the model to predict for the unseen slot-values, the copy mechanism is incorporated into this model.
Initially an attention score $a_i$ is calculated for each of the input tokens $x_i$ based on its hidden representation $h_i$ of the encoder.
We use a similar attention approach of \cite{Bahdanau2015} to compute the scores for each $h_i$.
Finally, for all tokens in a slot-value $v$, the corresponding attention scores in the input are combined to yield the copy score $\psi_c(v)$ for value $v$.
Formally, 
\begin{align}
    a_i &= tanh(W_c[S;h_i])\\
    \psi_c(v) &= \sum_t a_t \quad: x_t \in v
\end{align}
where $a_i$ is the attention score, $tanh$ is a non-linear activation function and $W_c$ is model parameter learned in training.

\subsubsection{Value score}
The value score $\psi_p(v)$ helps the model in predicting for the value $v$ based on the context and semantics of the source sentence.
The hidden representations $h_i$ of the encoder are weighted based on the normalized attention score $a$ to yield a context vector $C$.
Finally, the dot product of the context vector $C$ and the slot-value vectors $Z_s$ provides the value score for each value $v$.
Formally,
\begin{align}
    \alpha &= Softmax(a) \\
    C &= \sum_i \alpha_ih_i \\
    \psi_p &= C \cdot Z_s
\end{align}

Since $Z_s$ is computed using the slot values $V_s$, the \textit{value score} can be computed for any number for slot values.
This enables the model to update slot values when needed and predict for new values without the need for retraining it.

\section{Experiments}

\subsection{Datasets}
We experiment our model on the datasets of dialogue state tracking challenges (DSTC2, DSTC3) and Wizard-of-Oz (WoZ2.0).
The DSTC2 and WoZ2.0 are used to evaluate the models performance in predicting the seen slot values while the modified DSTC2 (explained in \ref{DSTC2}) and DSTC3 datasets are used to evaluate the models performance in predicting also the unseen slot values.

\subsubsection{DSTC2} \label{DSTC2}
The DSTC2 \cite{dstc2challenge} is a human-machine conversation dialogue dataset collected using Amazon Mechanical Turk consisting of dialogues in restaurants domain.
It is the standard benchmark dataset used for the task of dialogue state tracking.
DSTC2 consists of labels for tracking both informable slots and requestable slots at each turn.
We are interested in tracking the slot values for informable slots.
DSTC2 is a spoken dataset consisting of ASR hypotheses and turn-level semantic labels along with the transcriptions.
It consists of 1612 dialogues for training, 506 dialogues for development and 1117 dialogues for testing.
The possible values for slots in the ontology are the same for both train and testset.

Since DSTC2 dataset does not have any unseen slot values in the testset, we also create a modified version of this dataset using the same approach as in \cite{Xu2018}.
In particular, we randomly select $35\%$ of values (26 out of 74) for the slot \textit{food} and create a modified training data discarding any instance that contain one of the randomly selected value as the truth value for \textit{food} slot.
We use the top ASR hypotheses for both training and evaluation.

\subsubsection{DSTC3}
The DSTC3 \cite{dstc3challenge} dataset is a human-machine conversation dialogue dataset collected using Amazon Mechanical Turk consisting of dialogues in tourist domain.
DSTC3 was released to address the problems of unseen slot values and adaptability of a DST to a new domain.
It consists of 2265 dialogues for test with no specific dataset for train.
Since this dataset is an extension to the DSTC2 dataset, we use the DSTC2 dataset for training the model.
In particular, we combine the train and test sets of DSTC2 together yielding 2729 (1612+1117) training dialogues and use the same development set of DSTC2.
Though DSTC3 consists of new slots compared to DSTC2, we focus on predicting the unseen values for a given slot.
Therfore we predict only on the slots \textit{area}, \textit{food} and \textit{price} which are common in both DSTC2 and DSTC3.
The statistics of unseen values for each of these slot is shown in Table \ref{tab:dstc3_stat}.
We use the top ASR hypotheses for both training and evaluation.
\begin{table}
    \centering
    \begin{tabular}{c|ccc}
        \textbf{Slot} & \textbf{\#Values} & \textbf{Seen} & \textbf{Unseen} \\
        \hline
        Area & 20 & 6 & 14\\
        Food & 87 & 75 & 12 \\
        Price & 5 & 4 & 1\\
        \hline
        & 112 & 85 & 27
    \end{tabular}
    \caption{Statistics of seen and unseen values for each slot in DSTC3 dataset (excluding \textit{None} value).}
    \label{tab:dstc3_stat}
\end{table}

\subsubsection{WoZ2.0}
The WoZ2.0 \cite{NBT} dataset, collected using the Wizard of Oz framework, consists of written text conversations.
Each turn in the dialogue was contributed by different users who had to review all previous turns in that dialogue before contributing to the turn.
WoZ2.0 consists of a total of 1200 dialogues, out of which  600 for training, 200 for development and 400 for testing.

\subsection{Evaluation metrics}
We evaluate our model using the standard metrics for dialogue state tracking namely, \textit{accuracy} and \textit{joint goal}.
In particular, \textit{accuracy} is used to evaluate the model in tracking each slot separately, while \textit{joint goal} is used to evaluate the overall performance of the model in tracking the dialogue state.
\begin{enumerate}
    \item \textbf{Accuracy} is the ratio of number of turns where the slot is predicted correctly over the total number of turns and is calculated for each slot separately. We follow the \textit{Scheme A} and \textit{Schedule 1} evaluation scheme defined in the DSTC2 \cite{dstc2challenge} challenge.
    \item \textbf{Joint Goal} indicates the performance of the model in correctly tracking the goal constraints over a dialogue. The joint goal is the set of accumulated turn level goals up to a given turn.
\end{enumerate}

The objective of our work is to model a data-driven approach that is able to perform consistently in predicting both the seen and unseen slot-values.
Thus our goal is not to outperform any of the previous DST systems but rather address the weakness of existing system in predicting unseen slot values.

\subsection{Implementation}
We use pytorch\footnote{\url{https://pytorch.org/}} library to implement our model.
The encoder of the model is shared across all slots and a separate decoder is defined for each slot.
For the embeddings, we use the pre-trained Paragram-SL999 vectors \cite{wieting2015} of dimensions 300 learned using the Paraphrase Database (PPDB) \cite{ppdb}, and pre-trained character $n$-gram embeddings \cite{hashimoto2017} of dimension 100.
Both embeddings are concatenated resulting in an embedding of size 400 for each token and are fixed during training.
The number of hidden units in LSTM is set to 200 and a dropout of 0.2 is applied between different layers.

Since our DST is modeled as binary prediction for each value $v$, it can be used to track for domains where multiple values are possible for a single slot.
For the considered datasets, since a slot can take only a single value at a given turn, we use the top prediction.
The turn-level predictions are accumulated forward through the dialogue and the goal for slot $s$ is \textit{None} until it is predicted as value $v$ by the model.
The implemented model is experimented on 5 different random initializations for each dataset and the scores reported are the average of those experiments.

\subsection{Results}
\begin{table*}
    \centering
    \begin{tabular}{c|c|c}
        \multirow{2}{*}{\textbf{Model}}& \multicolumn{2}{c}{\textbf{Joint Goal}} \\
        & \textbf{DSTC2} & \textbf{WoZ2.0} \\
        \hline
        Delexicalisation-Based (DB) Model\cite{NBT} & \hspace{1mm} 69.1 & \hspace{1mm} 70.8 \\
        HDSTM \cite{Yang2018} & \hspace{1mm} 68.4 & \hspace{1mm} 84.5 \\
        Scalable multi-domain DST \cite{Rastogi2017} & \hspace{1mm} 70.3 & \hspace{1mm} - \\
        Pointer Net \cite{Xu2018} & \hspace{1mm} 72.1 & \hspace{1mm} - \\
        Neural Belief Tracker (NBT) - DNN \cite{NBT} & \hspace{1mm} 72.6 & \hspace{1mm} 84.4 \\
        Neural Belief Tracker (NBT) - CNN \cite{NBT} & \hspace{1mm} 73.4 & \hspace{1mm} 84.2 \\
        GLAD \cite{GLAD} & \hspace{1mm} 74.5 & \hspace{1mm} 88.1 \\
        StateNet\_PSI \cite{Ren2018} & \hspace{1mm} 75.5 & \hspace{1mm} 88.9 \\
        \hline  
        Our approach & \hspace{1mm} 73.8 & \hspace{1mm} 87.5\\
    \end{tabular}
    \caption{Dialogue state tracking performance on the testset of DSTC2 and WoZ2.0 datasets.}
    \label{tab:DSTC2_WOZ}
\end{table*}
The joint goal performance of the model on both the DSTC2 dataset and WoZ2.0 dataset is shown in Table \ref{tab:DSTC2_WOZ}.
The delexicalisation-based (DB) model, as reported in \cite{NBT}, is based on \cite{Henderson2015} for the DSTC2 dataset and on \cite{Wen2017} for the WoZ2.0 dataset.
The DB model relies on a semantic lexicon to replace slot names in the input to generic delexicalised token while all other approaches in Table \ref{tab:DSTC2_WOZ} (including our approach) are data-driven.
The approaches of neural belief tracker (NBT) \cite{NBT}, GLAD \cite{GLAD} and StateNet\_PSI \cite{Ren2018} are focused on modelling the seen slot-values, while HDSTM \cite{Yang2018}, multi-domain DST \cite{Rastogi2017} and Pointer Net \cite{Xu2018} addresses the problem of unseen slot values in some aspect.
We can see from the results that our approach significantly improves on the NBT and Pointer Net on both DSTC2 and WoZ2.0 datasets proving its ability in performing state-of-the-art in predicting the seen slot-values.

Though StateNet\_PSI model claim to be able to predict for unseen slot-values, they force a hard constraint on the presence of a pretrained semantic embedding for any unseen slot-values.
This is not ideal for slots such as \textit{area}, \textit{food} or \textit{location} which usually contain names that do not have pretrained embedding.
Our approach is flexible in this aspect since the copy mechanism can point the relevant token in the input and make appropriate prediction without requiring pretrained embedding.
The GLAD model on the other hand uses an encoder with LSTM nodes to represent the slot-values which makes it unsuitable to effectively predict for unseen slot-values.

\begin{figure}
    \centering
    \includegraphics[width=\columnwidth]{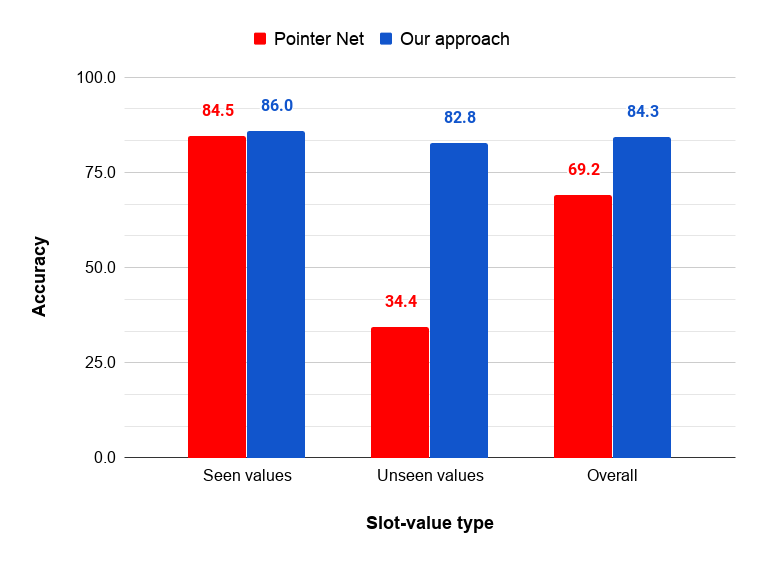}
    \caption{Accuracy of food slot on modified DSTC2 dataset.}
    \label{fig:result_comparison}
\end{figure}
We show the performance of our model in predicting unseen slot values on the modified DSTC2 dataset in Figure \ref{fig:result_comparison}.
We can see that the Pointer Net \cite{Xu2018} approach though performs well on seen slot-values, drops in performance on unseen slot-values.
This could be due to the hybrid architecture employed in Pointer Net where prediction of a slot value is done in two stages (i.e predict if the value is \textit{None} or \textit{dontcare} or \textit{other} and if \textit{other}, then predict the location of slot value in the input source).
Since this approach points to the location of slot value in the input source, the training data should contain the location of the reference slot value and it also requires an additional post-processing step to normalize for the value.
Our approach does not provide any such constraints and shows a consistent performance on both seen and unseen slot-values.

\begin{table*}
    \centering
    \begin{tabular}{c|ccc|ccc}
        \multirow{2}{*}{\textbf{Model}} & \multicolumn{3}{c|}{\textbf{DSTC2}} & \multicolumn{3}{c}{\textbf{DSTC3}} \\
        & \textbf{Area} & \textbf{Food} & \textbf{Price} & \textbf{Area} & \textbf{Food} & \textbf{Price} \\
        \hline
        Focus Baseline & 90.8 & 83.7 & 92.9 & 81.1 & 90.5 & 88.4 \\
        RNN with rules & 92.4 & 85.6 & 93.0 & 88.5 & 91.0 & 93.2 \\
        \hline
        Cos with Sentinel \cite{yoshida2018} & 84.7 & 84.4 & 83.7 & 80.6 & 79.6 & 66.9 \\
        Our approach & 91.0 & 81.5 & 92.4 & 78.0 & 84.0 & 91.5\\
    \end{tabular}
    \caption{Overall accuracy of the model on both DSTC2 and DSTC3 datasets.}
    \label{tab:acc_ov_res}
\end{table*}
\begin{table}
    \centering
    \begin{tabular}{c|ccc}
        \multirow{2}{*}{\textbf{Model}} & \multicolumn{3}{c}{\textbf{DSTC3}} \\
        & \textbf{Area} & \textbf{Food} & \textbf{Price} \\
        \hline
        Focus Baseline & 67.8 & 88.1 & 87.6\\
        RNN with rules & 85.3 & 82.3 & 92.3\\
        \hline
        Cos with Sentinel & \multirow{2}{*}{71.5} & \multirow{2}{*}{59.5} & \multirow{2}{*}{52.7}\\
        \cite{yoshida2018} & & &\\
        Our approach & 68.5 & 77.9 & 85.0 \\

    \end{tabular}
    \caption{Accuracy on model in predicting the unseen slot values in DSTC3 dataset.}    
    \label{tab:acc_us_res}
\end{table}
The overall slot-wise accuracy of our model on the DSTC2 and DSTC3 datasets are shown in Table \ref{tab:acc_ov_res}.
The DSTC2 dataset consists only of seen slot values while DSTC3 has both seen and unseen slot-values.
The \textit{Focus Baseline} and \textit{RNN with rules} are the results from the official DSTC2 and DSTC3 challenge \cite{dstc2challenge, dstc3challenge}.
The \textit{focus baseline} is a rule based approach that uses SLU results to track the dialogue state while \textit{RNN with rules} uses recurrent neural network using delexicalisation approach.
\textit{RNN with rules} showed the best accuracy in DSTC3 in the official challenge.
Since these two models require external features such as SLU or a semantic lexicon, their scalability to real world applications are very poor and require significant feature modelling for each domain.
The \textit{Cos with Sentinel} \cite{yoshida2018} is a data-driven approach which uses cosine similarity between the slot value and the inputs, and makes prediction using a sentinel mixture model.
We can see that a complete data-driven approach such as \textit{Cos with sentinel}, that is modelled to predict both seen slots and unseen slot values suffers from lower performance also on seen slot values.
However our approach is able to achieve better performance on DSTC3 dataset (consisting unseen slot-values) while performing close to state-of-the-art in DSTC2 dataset.
The performance of the models on predicting only the unseen slot values in DSTC3 dataset in shown in Table \ref{tab:acc_us_res}.

\subsection{Discussion}
From the results shown in Tables \ref{tab:DSTC2_WOZ}, \ref{tab:acc_ov_res} and \ref{tab:acc_us_res}, we can infer that using copy mechanism for the task of dialogue state tracking provides a more robust model.
While previous approaches either rely on semantic lexicon, delexicalization or rules to predict the unseen slot-values, our approach takes a complete data-driven path to address this problem.
Though we use semantic embeddings as part of our word embedding, our model is also able to handle tokens for which no semantic embedding exist. This can be seen in the accuracy performance of \textit{area} and \textit{food} slots in DSTC3 dataset, where the unseen slot values are typically proper names for which the semantic embedding did not exist.
The copy mechanism combined with the $n$-gram embedding helps the model in handling such scenarios.
While our approach is used to predict from on a pre-defined set (categorical) of slot-values in this work, it can also be adopted to scenarios where the slot-value is unbounded.
This provides advantage in modelling for a schema-based DST, where the datatype of a slot can either be categorical or free-form.

\section{Conclusion}
In this paper, we addressed the problem of unseen slot values for the task of dialogue state tracking and proposed a novel E2E data-driven approach based on copying mechanism that is robust in tracking for both seen and unseen slot values.
The proposed approach offers advantage in being flexible for slot value updates in the ontology without relying on either SLU or handcrafted features.
Our evaluations demonstrate that the proposed approach provides considerate advantage in being able to track unseen slot values effectively with state-of-the-art performance.
In future works we would like to investigate our approach targeting the applicability of the model in multi-domain dialogue state tracking.

\bibliography{conll-2019}
\bibliographystyle{acl_natbib}

\end{document}